# LiDAR-based SLAM for robotic mapping: state of the art and new frontiers

Xiangdi Yue and Miaolei He*

College of Engineering and Design, Hunan Normal University, Changsha 410081, China.

Yihuan Zhang

Intelligent Connected Vehicle Center, Tsinghua Automotive Research Institute, Suzhou 215000, China.

*Corresponding author. E-mail: rchml@hotmail.com.

**Abstract**

**Purpose** - In recent decades, the field of robotic mapping has witnessed widespread research and development in LiDAR (Light Detection And Ranging)-based simultaneous localization and mapping (SLAM) techniques. This paper intend to provide a significant reference for researchers and engineers in robotic mapping.

**Design/methodology/approach -** This paper focused on the research state of LiDAR-based SLAM for robotic mapping as well as a literature survey from the perspective of various LiDAR types and configurations.

**Findings -** This paper conducted a comprehensive literature review of the LiDAR-based SLAM system based on three distinct LiDAR forms and configurations. We concluded that multi-robot collaborative mapping and multi-source fusion SLAM systems based on 3D LiDAR with deep learning will be new trends in the future.

**Originality/value –** As far as the authors' knowledge permits, this is the first thorough survey of robotic mapping from the perspective of various LiDAR types and configurations. It can serve as a theoretical and practical guide for the advancement of academic and industrial robot mapping.

## 1. Introduction

With the rapid advancements in artificial intelligence technology, mobile robots have increasingly taken on the role of human operators in various practical operations, offering improved efficiency and safety. Consequently, these robotic systems, encompassing sensors, remote controls, automatic controllers, and other mobile capabilities, have become integral components in an array of application scenarios. State estimation and localization in unknown environments have emerged as prominent research areas in the domain of mobile robotics, with SLAM serving as a focal point. Compared to cameras, the utilization of LiDAR technology provides notable advantages, as it is unaffected by ambient light and texture, allowing for highly accurate and efficient distance measurements. The LiDAR-based SLAM system has been extensively developed in the fields of automated driving (Y. Zhang, *et al.*, 2022, C. Badue, *et al.*, 2021), mobile robots, forestry surveying (S. Tao, *et al.*, 2021), urban surveying and mapping (L. Liu, *et al.*, 2017).

Tee provided a comprehensive analysis and comparison of several popular open-source implementations of 2D LiDAR-based SLAM (Y.K. Tee and Y.C. Han, 2021). However, the investigation solely focuses on 2D LiDAR-based SLAM techniques, with no mention of their 3D counterparts. Bresson examined the application of LiDAR-based SLAM specifically within the context of the grand challenge of autonomous driving (G. Bresson, *et al.*, 2017). Notably, Xu presented an in-depth exploration of the development of multi-sensor fusion positioning, with meticulous attention given to the evaluation of both loosely coupled and tightly coupled systems (X. Xu, *et al.*, 2022). This paper presents a novel approach to reviewing the literature on LiDAR-based SLAM by focusing on the application of different types and configurations of LiDAR. This paper offers a significant contribution as a reference for researchers and engineers seeking to gain insight into the wide-ranging applications of different LiDAR types and configurations, distinguishing itself from previous review studies.

The remainder of this paper is organized as follows: Section 2 provides an anatomy of a LiDAR-based SLAM system. In Section 3, the related work of LiDAR-based SLAM systems is reviewed in three segments based on LiDAR types and configurations. Section 4 proposes several new frontiers in LiDAR-based SLAM. Finally, Section 5 concludes this paper.

## 2. Anatomy of a LiDAR-based SLAM system

### 2.1 Historical perspective

Smith and Cheeseman united the robot localization and mapping problems in 1986 under a theoretical framework based on probability theory, which can be considered the beginning of SLAM problem study (R.C. Smith and P. Cheeseman, 1986). In 2006, Durrant-Whyte and Bailey used the term SLAM for the first time in their research papers (H. Durrant-Whyte and T. Bailey, 2006). This was also known as the SLAM classical period. The development history of LiDAR-based SLAM is shown in Figure 1. Development history of LiDAR-based SLAM.

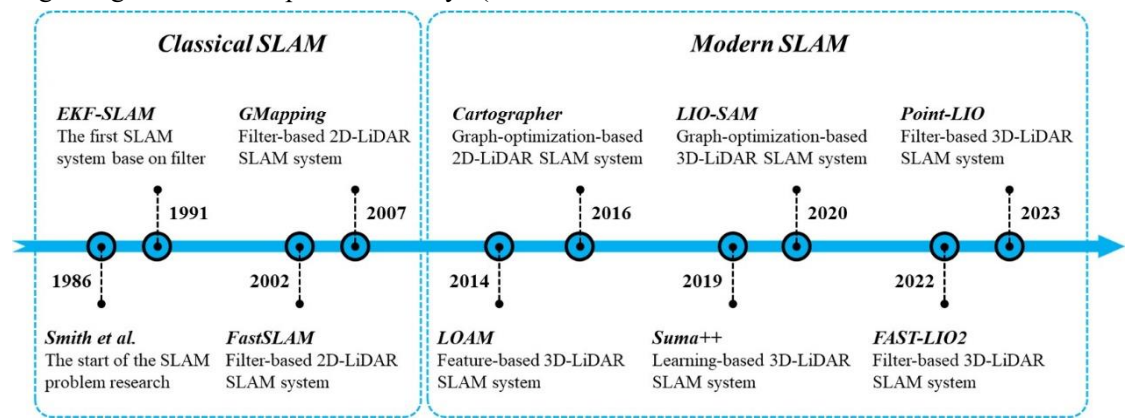

Figure 1. Development history of LiDAR-based SLAM

The filtering approach was the primary way used to tackle the SLAM problem throughout the classical period. To solve SLAM in a Bayesian network, the filtering algorithm must gather information at each moment in real time and partition it into the Bayesian network's probability distribution. This filtering method represents an online SLAM system, which, as is evident, incurs significant computational overhead and can only generate maps on a small scale. For large-scale mapping, an optimization strategy for solving SLAM in factor graphs has been proposed. The optimization method is the inverse of the filtering method, which merely accumulates acquired information and calculates offline the robot's trajectory and waypoints using the global information accumulated at all previous instances. In other terms, the method for optimization is a complete SLAM system. With the tremendous increase in computer performance and mathematical capability, optimization-based approaches have become the primary focus of contemporary SLAM research. Consequently, Section 2 analyzes the SLAM framework based on the optimization method in a contemporary manner (shown in Figure 2. LiDAR-based SLAM system overview).

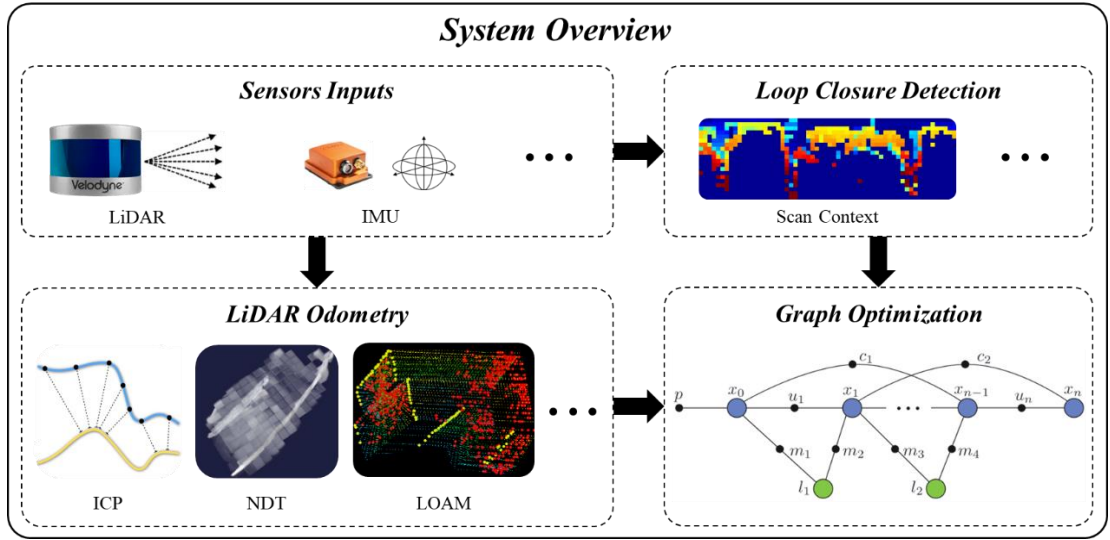

Figure 2. LiDAR-based SLAM system overview

### 2.2 LiDAR odometry

The purpose of the LiDAR odometry is to produce a local map by creating an estimate of the motion between two neighboring point cloud frames. LiDAR odometry is classified into three types depending on point cloud registration methods: point-based registration, distribution-based registration, and feature-based registration.

Point-based point cloud registration finds the correspondence between the target and reference point clouds in the most straightforward way possible. A simple way to identify the corresponding points in the reference point cloud is to find the one with the shortest Euclidean distance, i.e., the closest point. Besl described a general-purpose, representation-independent method for the accurate and computationally efficient registration of 3-D shapes based on the iterative closest point (ICP) (P.J.B.a.N.D. McKay, 1992) algorithm, which determines the optimum Euclidean transformation with the fewest square distances from point-to-point correspondences.

Instead of points, the distribution-based registration method converts the point cloud space into voxels with a continuous probability density function. Matching the continuous probability density function of the target point cloud with the reference point cloud optimises the pose connection. In 2003, Biber first demonstrated an alternative representation for a range scan, which is called the normal distributions transform (NDT) (P. Biber and W. Strasser, 2003). Similar to an occupancy grid, Biber subdivided the 2D plane into cells. For each cell, an anormal distribution was assigned, which locally modeled the probability of measuring a point.

ICP and NDT concepts are based on the direct registration of point clouds. This approach of direct registration is time-consuming and difficult to provide in real-time. Consequently, Zhang presented LOAM (J. Zhang and S. Singh, 2014) to execute point cloud registration of two neighboring frames by extracting geometric features, which increased the system's efficiency and enhanced registration accuracy.

### 2.3 Loop closure detection

Global data association corrects cumulative mistakes by recognising if the robot has reached the location it has arrived at in the historical instant to produce a globally consistent map. Loop closure detection usually has two steps: (1) Use position recognition to find a point in the database similar to the current observation. (2) Pose graph optimization corrects the estimated loop posture. The method for detecting loop closures using LiDAR may be further categorised into two approaches: local-descriptor-based and global-descriptor-based.

### 2.4 Graph optimization

The accumulated errors of LiDAR make the map inaccurate for a long period, yet the LiDAR odometry can produce a trajectory and map quickly. Thus, to determine the ideal route and map for a long period using LiDAR odometry, a large-scale optimization problem must be created. Graph optimization is a method for achieving overall optimization by combining the pose and inter-frame motion constraints of each radar frame. It assists in eliminating local accumulated errors in large-scale mapping and coordinates the previous trajectory.

## 3. LiDAR-based SLAM systems for robotic mapping

In this section, we will conduct a comprehensive literature review of the LiDAR-based SLAM system based on three distinct LiDAR forms and configurations, including (1) 2D LiDAR-based SLAM systems; (2) 3D LiDAR-based SLAM systems; and (3) spinning-actuated LiDAR-based SLAM systems.

### 3.1 2D LiDAR-based SLAM systems

A single-line LiDAR is comprised of a single-line laser module and a rotating mechanism. The scanning point of single-line LiDAR is typically within a 360-degree range on the same plane, i.e., the contour of a particular cross-section of the environment. Hence, it is also known as 2D LiDAR. 2D LiDAR-based SLAM is a top-view LiDAR SLAM algorithm compared to the three-dimensional point cloud. This simplifies laser scanning and maps the data into two dimensions. They resemble images. Two-dimensional SLAM can use image feature extraction and matching algorithms to save the map as a picture. Indoor sweepers, service robots, and AGVs mostly use 2D LiDAR-based SLAM. Next, we will discuss the state-of-the-art of 2D LiDAR-based SLAM algorithms nowadays.

In 2002, Montemerlo proposed the FastSLAM (M. Montemerlo, *et al.*, 2002) algorithm, which used particle filters and Kalman filters to estimate robot posture and position landmarks, respectively. The GMapping (G. Grisetti, *et al.*, 2007) algorithm framework is founded on the RBPF (Rao-Blackwellized Particle Filter) algorithm, which initially locates and then maps. Konolige developed the Karto (K. Konolige, *et al.*, 2010) SLAM algorithm in 2010, which was the first open-source graph optimization algorithm. To solve sparse decoupling, it employed height direction optimization and non-iterative square root decomposition. But this algorithm must build a local sub-map beforehand in the loop closure detection section. The framework of the Hector (S. Kohlbrecher, *et al.*, 2011) SLAM algorithm was founded on Gauss-Newton. The algorithm was free of an odometer and appropriate for aerial or uneven road conditions. However, when the robot turned rapidly, it was prone to error matching, and there was no loop closure detection module. In 2016, Google introduced Cartographer (W. Hess, *et al.*, 2016), a sensor-equipped knapsack that generated 2D grid maps with a resolution of r = 5 cm in real time for indoor mapping. This algorithm completed the front-end matching with correlation scanning combined with gradient optimization and used the depth-first branch and bound search algorithm to calculate loop closure detection. Macenski built a set of tools and capabilities for 2D SLAM called SLAM Toolbox (S. Macenski and I. Jambrecic, 2021). SLAM Toolbox offered synchronous and asynchronous mapping modes, localization, multi-session mapping, graph optimization, reduced compute time, and prototype lifelong and distributed mapping applications.

Table 1 are summarized to present the worldwide research status of 2D LiDAR-based SLAM systems.

Table 1. The state-of-the-art of 2D LiDAR-based SLAM systems

| System | Sensor | Framework | Source |
|---|---|---|---|
| FastSLAM | 2D LiDAR | Filter | (M. Montemerlo, S. Thrun, D. Koller and B. Wegbreit, 2002) |
| GMapping | 2D LiDAR | Filter | (G. Grisetti, C. Stachniss and W. Burgard, 2007) |
| Karto SLAM | 2D LiDAR | Graph optimization | (K. Konolige, G. Grisetti, R. Kümmerle, W. Burgard, B. Limketkai and R. Vincent, 2010) |
| Hector SLAM | 2D LiDAR | Gauss-Newton | (S. Kohlbrecher, O. Von Stryk, J. Meyer and U. Klingauf, 2011) |
| Cartographer | 2D LiDAR and IMU | Graph optimization | (W. Hess, D. Kohler, H. Rapp and D. Andor, 2016) |
| SLAM Toolbox | 2D LiDAR | / | (S. Macenski and I. Jambrecic, 2021) |

### 3.2 3D LiDAR-based SLAM systems

2D LiDAR can only scan obstacle information on the same plane, i.e., the contour of a cross section of the environment, so the information extracted from the scan is extremely limited. Multi-line LiDAR, also known as 3D LiDAR, enables you to scan the contours of multiple cross-sections by simultaneously emitting multiple laser beams in the vertical direction in conjunction with a rotating mechanism. 3D LiDAR-based SLAM is widely used in the field of outdoor mobile robotics and autonomous driving due to its ability to provide rich point cloud information about the surrounding environment. According to different frameworks, the field of 3D LiDAR-based SLAM can be further categorized into two distinct schemes: filter-based and graph optimization-based. Subsequently, we will go over the state-of-the-art of 3D LiDAR-based SLAM algorithms. Table 2 are summarized to present the worldwide research status of 3D LiDAR-based SLAM systems.

Table 2. The state-of-the-art of 3D LiDAR-based SLAM systems

| System | Sensor | Framework | Source |
|---|---|---|---|
| LeGO-LOAM | 3D LiDAR and IMU | Graph optimization | (T. Shan and B. Englot, 2018) |
| SuMa | 3D LiDAR | Graph optimization | (J. Behley and C. Stachniss, 2018) |
| hdl-graph-slam | 3D LiDAR | Graph optimization | (K. Koide, *et al.*, 2019) |
| LINS | 3D LiDAR and IMU | Filter | (C. Qin, *et al.*, 2020) |
| LIO-SAM | 3D LiDAR and IMU | Graph optimization | (T. Shan, *et al.*, 2020) |
| FAST-LIO | 3D LiDAR and IMU | Filter | (W. Xu and F. Zhang, 2021) |
| BALM | 3D LiDAR | Graph optimization | (Z. Liu and F. Zhang, 2021) |

| F-LOAM | 3D LiDAR | Graph optimization | (H. Wang, et al., 2021) |
| E-LOAM | 3D LiDAR | Graph optimization | (H. Guo, et al., 2022) |
| D-LIOM | 3D LiDAR and IMU | Graph optimization | (Z. Wang, et al., 2022) |
| ART-SLAM | 3D LiDAR and IMU | Graph optimization | (M. Frosi and M. Matteucci, 2022) |
| LOCUS 2.0 | 3D LiDAR and IMU | Graph optimization | (A. Reinke, et al., 2022) |
| FAST-LIO2 | 3D LiDAR and IMU | Filter | (W. Xu, et al., 2022) |
| Faster-LIO | 3D LiDAR and IMU | Filter | (C. Bai, et al., 2022) |
| EKF-LOAM | 3D LiDAR and IMU | Filter | (G.P.C. Junior, et al., 2022) |
| VoxelMap | 3D LiDAR and IMU | Filter | (C. Yuan, et al., 2022) |
| Point-LIO | 3D LiDAR and IMU | Filter | (D. He, et al., 2023) |
| Inv-LIO1 | 3D LiDAR and IMU | Filter | (P. Shi, et al., 2023) |

**3.2.1 Filter-based SLAM systems**

Huang investigated the observability of the consistency of extended Kalman filter (EKF)-based cooperative localization (CL) (G.P. Huang, et al., 2011). Analytically, he demonstrated that the error-state system model employed in the standard EKF-based CL always had a larger observable subspace than the actual nonlinear CL system. LINS (C. Qin, H. Ye, C.E. Pranata, J. Han, S. Zhang and M. Liu, 2020) was a lightweight LiDAR-inertial state estimator for ego-motion estimation in real-time. By tightly coupling a 6-axis IMU and a 3D LiDAR, it enabled robust and efficient ground vehicle navigation in challenging environments, such as featureless scenarios. In this system, an iterated error-state Kalman filter (iESKF) was developed to repeatedly correct the approximated state by generating new feature correspondences with each iteration, while keeping the system computationally accessible. Xu of the University of Hong Kong proposed FAST-LIO2 (W. Xu, Y. Cai, D. He, J. Lin and F. Zhang, 2022), which was based on the second generation of FAST-LIO (W. Xu and F. Zhang, 2021), which was also based on the iESKF. In terms of the design of the filter, FAST-LIO2 and LINS were comparable, but the calculation of Kalman gain differed. Soon after, Gao developed Faster-LIO (C. Bai, T. Xiao, Y. Chen, H. Wang, F. Zhang and X. Gao, 2022) based on FAST-LIO2. This algorithm's advantage over FAST-LIO2 was that it achieved greater algorithmic efficiency while maintaining accuracy. This was primarily due to the fact that the iVox (incremental Voxels) data structure was used to maintain the local map, which could effectively reduce the point cloud registration time without influencing the odometer's accuracy. This work is currently compatible with both mechanical and solid-state LiDAR. Gilmar presented EKF-LOAM (G.P.C. Junior, A.M.C. Rezende, V.R.F. Miranda, R. Fernandes, H. Azpurua, A.A. Neto, G. Pessin and G.M. Freitas, 2022), an enhanced 3D LiDAR-based SLAM strategy that incorporated wheel odometry and the IMU into the SLAM process. Yuan provided VoxelMap (C. Yuan, W. Xu, X. Liu, X. Hong and F. Zhang, 2022), an efficient and probabilistic adaptive voxel mapping method for LiDAR odometry. The map was comprised of voxels, each of which contained a single plane feature that allowed for the probabilistic representation of the environment and the precise registration of a new LiDAR scan. Point-LIO (D. He, W. Xu, N. Chen, F. Kong, C. Yuan and F. Zhang, 2023), a robust and high-bandwidth light detection and ranging (LiDAR) inertial odometry with the capability to estimate extremely aggressive robotic motions. Shi presented Inv-LIO1 (P. Shi, Z. Zhu, S. Sun, X. Zhao and M. Tan, 2023), a robo-centric invariant EKF LiDAR-inertial odometry. This system directly fused LiDAR and IMU measurements using invariant observer design and the theory of Lie groups.

### 3.2.2 Graph optimization-based SLAM systems

Shan introduced the LeGO-LOAM (T. Shan and B. Englot, 2018) algorithm for real-time six-degree-of-freedom ground vehicle posture estimation. Jens constructed a surfel-based map and estimated the pose changes of the robot by exploiting the projective data association between the present scan and a rendered model view from the surfel map (J. Behley and C. Stachniss, 2018). Koide described a three-dimensional LiDAR-based portable people-behaviour measuring system (K. Koide, J. Miura and E. Menegatti, 2019). The system tracked the target person while estimating the sensor's pose in a three-dimensional ambient map. Shan proposed the LIO-SAM (T. Shan, B. Englot, D. Meyers, W. Wang, C. Ratti and D. Rus, 2020), which was based on LeGO-LOAM, by tightly coupling LiDAR and IMU. LIO-SAM only used the sliding window to optimise the IMU's deviation. Then, it used an additional back-end to put the IMU pre-integration factor, the LiDAR odometer factor, the GPS factor, and the loop closure detection factor into a factor graph optimization model for joint optimization in order to obtain the robot's globally consistent pose. A local Bundle Adjustment (BA) on a sliding window of keyframes has been frequently utilised in visual SLAM to reduce drift. Hence, Liu formulated the LiDAR BA as minimising the distance from a feature point to its matched edge or plane (Z. Liu and F. Zhang, 2021). This method could greatly reduce the optimization scale and allow large-scale dense plane and edge features to be used. Wang suggested F-LOAM (H. Wang, C. Wang, C.-L. Chen and L. Xie, 2021), a non-iterative two-stage distortion compensation method, to reduce computational cost and provide a computationally efficient and accurate framework for LiDAR-based SLAM. Guo proposed E-LOAM (LOAM with Expanded Local Structural Information) (H. Guo, J. Zhu and Y. Chen, 2022), which added local point cloud information around geometric feature points to pre-extracted geometric information. Wang introduced D-LIOM (Z. Wang, L. Zhang, Y. Shen and Y. Zhou, 2022), a tightly coupled direct LiDAR-inertial odometry and mapping architecture. D-LIOM immediately registered a scan to a probability submap and integrated LiDAR odometry, IMU pre-integration, and gravity constraint to generate a local factor graph in the submap's time window for real-time high-precision pose estimation. ART-SLAM (Accurate Real-Time LiDAR SLAM) (M. Frosi and M. Matteucci, 2022) was a modular, fast, and accurate LiDAR SLAM system for batch and online estimation. Using a three-phased algorithm, this system was able to efficiently detect and close loops in the trajectory. Andrzej presented LOCUS 2.0 (A. Reinke, M. Palieri, B. Morrell, Y. Chang, K. Ebadi, L. Carlone and A.-a. Agha-mohammadi, 2022), a robust and computationally-efficient LiDAR odometry system for real-time underground 3D mapping.

### 3.3 Spinning-actuated LiDAR-based SLAM systems

The mechanical LiDAR horizontal FOV (field of view) is 360°, but the vertical FOV is limited. Solid-state LiDAR has a broad vertical field of view but a small horizontal field. Existing research usually uses a multi-LiDAR scheme (J. Jiao, *et al.*, 2022, M. Velas, *et al.*, 2019) or a spinning-actuated-LiDAR method to enhance LiDAR's field of view, but the former is too expensive, so the latter is the dominant trend. Spinning-actuated LiDAR-based SLAM systems are primarily utilized for applications requiring panoramic scanning coverage, including surveying and mapping (T. Lowe, *et al.*, 2021), subterranean exploration (E. Jones, *et al.*, 2020), etc. Subsequently, we will discuss the state-of-the-art of spinning-actuated LiDAR-based SLAM algorithms.

Zhen proposed a unified mapping framework (UMF) (W. Zhen and S. Scherer, 2020) that supported numerous LiDAR types, including (1) a fixed 3D LiDAR and (2) a rotating 3D/2D LiDAR. The localization module utilised a combination of an error state Kalman filter (ESKF) and a Gaussian particle Filter (GPF) to estimate robot states within the prior map. Mojtaba presented LoLa-SLAM (M. Karimi, *et al.*, 2021), a framework for low-latency LiDAR SLAM based on LiDAR scan segmentation and concurrent matching. This framework employed segmented point cloud data from a spinning-actuated LiDAR in a concurrent multithreaded matching pipeline to estimate 6D pose with a high update rate and low latency. Chen developed R-LIO (K. Chen, *et al.*, 2022) (rotating LiDAR inertial odometry), a novel SLAM algorithm that integrated a spinning-actuated 3D LiDAR with an IMU. R-LIO was capable of high-precision, real-time position estimation and map construction. Milad introduced Wildcat (L.C. de Lima, *et al.*, 2023), an elastic and robust online 3D LiDAR-based SLAM system. Wildcat's core used a continuous-time trajectory representation and an efficient pose-graph optimization module that supported single- and multi-agent scenarios. Chanoh presented a novel spinning-actuated 3D LiDAR-based map-centric SLAM framework (C. Park, *et al.*, 2022). Possessing the benefits of a map-centric approach, this method exhibited novel characteristics to overcome the deficiencies of existing systems associated with multi-modal sensor fusion and LiDAR motion distortion. Wang presented the online multiple calibration inertial odometer (OMC-SLIO) (S. Wang, *et al.*, 2022) approach for SLiDAR (spinning LiDAR), which estimated numerous extrinsic parameters of the LiDAR, rotating mechanism, IMU, and odometer state online. Yan introduced Spin-LOAM (L. Yan, *et al.*, 2023), a tightly coupled 3D LiDAR-based SLAM algorithm for spinning-actuated LiDAR systems. There was an adaptive scan accumulation method that analyzed feature point spatial distribution to increase matching accuracy and reliability. Chen demonstrated the powered-flying ultra-underactuated LiDAR sensing aerial robot (PULSAR) (N. Chen, *et al.*, 2023), a self-rotating, flexible UAV (Unmanned Aerial Vehicle) whose three-dimensional position was entirely controlled by actuating just one motor to produce the necessary thrust and moment. Table 3 are summarized to present the worldwide research status of spinning-actuated LiDAR-based SLAM systems.

Table 3. The state-of-the-art of spinning-actuated LiDAR-based SLAM systems

| System | Sensor | Framework | Source |
| --- | --- | --- | --- |
| UMF | 3D LiDAR and IMU | Filter | (W. Zhen and S. Scherer, 2020) |
| LoLa-SLAM | 3D LiDAR | Filter | (M. Karimi, M. Oelsch, O. Stengel, E. Babaians and E. Steinbach, 2021) |
| R-LIO | 3D LiDAR and IMU | Graph optimization | (K. Chen, K. Zhan, F. Pang, X. Yang and D. Zhang, 2022) |
| Wildcat | 3D LiDAR and IMU | Graph optimization | (L.C. de Lima, M. Ramezani, P. Borges and M. Brunig, 2023) |
| Map-centric SLAM | 3D LiDAR and IMU | Graph optimization | (C. Park, P. Moghadam, J. Williams, S. Kim, S. Sridharan and C. Fookes, 2022) |
| OMC-SLIO | 3D LiDAR and IMU | Filter | (S. Wang, H. Zhang and G. Wang, 2022) |
| Spin-LOAM | 3D LiDAR and IMU | Graph optimization | (L. Yan, J. Dai, Y. Zhao and C. Chen, 2023) |

## 4. Challenges

The quality of the mapping will have an immediate impact on subsequent high-order duties, such as decision-making and planning. Among them, LiDAR-based SLAM for mapping is a mature technology that has been extensively researched. While LiDAR-based SLAM-related work has made significant strides in the past few decades, there are still numerous challenges and issues that need to be resolved. In retrospect of the recent LiDAR-based SLAM towards robotic mapping, several aspects of the challenges and prospective research directions in robotic mapping will be discussed in the following.

- *LiDAR-based SLAM in degenerated environments.* Tunnels, bridges, and long corridors are typical degraded environments with no geometric elements, identical environments, and symmetrical architecture. The LiDAR-based SLAM system cannot estimate the full robot's 6-DOF motion in a degraded environment. LiDAR-based SLAM in degenerated Environments represents a significant challenge (H. Li, *et al.*, 2022, J. Jiao, *et al.*, 2021).
- *LiDAR-based lifelong SLAM in dynamic environments.* All LiDAR-based SLAM technologies operate under the assumption that the environment is static. Under dynamic environmental conditions, such as when a robot is creating a map, an object exists somewhere. When the robot is positioned using a prior map, the object is absent, resulting in the failure of autonomous positioning. Lifelong mapping utilising LiDAR can solve the problem of mapping in a dynamic environment (G. Kim and A. Kim, 2022, S. Zhu, *et al.*, 2021, M. Zhao, *et al.*, 2021).
- *LiDAR-based SLAM for large-scale environments.* Faced with large-scale environmental mapping needs, multi-robot cooperative mapping using LiDAR-based SLAM schemes can address the computational load, global error accumulation, and risk concentration issues that plague single-robot SLAM (Y. Xie, *et al.*, 2022, Y. Chang, *et al.*, 2022, H. Mahboob, *et al.*, 2023, P. Huang, *et al.*, 2021).
- *Multi-source fusion-enhanced LiDAR-based SLAM.* Multi-source fusion SLAM systems based on 3D LiDAR are another research hotspot. Considering that single sensors like LiDAR, camera, and IMU are inaccurate and fragile, researchers have increasingly developed multi-source fusion SLAM solutions (T. Shan, *et al.*, 2021, H. Tang, *et al.*, 2023, R. Lin, *et al.*, 2021, C. Zheng, *et al.*, 2022).
- *Deep learning-augmented LiDAR-based semantic SLAM.* Extensive research has been conducted on LiDAR-based SLAM systems enhanced with deep learning. Deep learning combined with LiDAR-based SLAM in robotic mapping will also be a potential research trend in the future. High-level semantic information-assisted LiDAR-based SLAM has become an essential tool in robotic mapping (X. Chen, *et al.*, 2019, S.W. Chen, *et al.*, 2020, S. Du, *et al.*, 2021).

## 5. Conclusions

This paper focused on the research state of LiDAR-based SLAM for robotic mapping from the perspective of various LiDAR types and configurations.

Initially, from a historical perspective, we go over the origin of SLAM. A framework for a modern SLAM system based on optimization methods is proposed by comparing and analyzing the characteristics of classical SLAM based on filtering methods. Subsequently, this paper undertakes an extensive literature review of the LiDAR-based SLAM system in three separate LiDAR forms and configurations. Compared to the three-dimensional point cloud, 2D LiDAR-based SLAM is a top-view LiDAR SLAM method. Most indoor sweeping robots, service robots, and AGVs use SLAM that is based on 2D-LiDAR. Multi-line LiDAR, also referred to as 3D LiDAR, allows you to scan the contours of several cross-sections. Nowadays, 3D LiDAR-based SLAM is widely applied in the fields of outdoor mobile robotics and autonomous driving. In general, spinning-actuated LiDAR-based SLAM systems are mostly used for tasks that need wide-angle scanning coverage, such as surveying and mapping, subterranean exploration, etc. Ultimately, challenges in LiDAR-based SLAM for robotic mapping are also briefly discussed. Multi-robot collaborative mapping and multi-source fusion SLAM systems based on 3D LiDAR with deep learning will be new trends in the future.

**Acknowledgment**

This work was supported by the Natural Science Foundation of Hunan Province, China, 2021JJ40353; the National Innovation and Entrepreneurship Training Program of China, 202210542043; the Natural Science Foundation of Zhejiang Province, China, LQ23E050015.